\documentclass[conference]{IEEEtran}
\IEEEoverridecommandlockouts

\usepackage{cite}

\usepackage{amsfonts}       
\usepackage{amssymb,amsmath}
\usepackage{nicefrac}       
\usepackage{microtype}      
\usepackage{algorithmic}
\usepackage{graphicx}
\usepackage{textcomp}
\usepackage[table]{xcolor}
\usepackage{comment}
\usepackage{caption}
\usepackage{makecell} 
\usepackage{multirow}
\usepackage{todonotes}
\usepackage{flushend}
\usepackage{mathtools}
\usepackage{pdflscape}
\usepackage{multicol}

\usepackage{tikz} 
\usepackage{pifont}

\def\BibTeX{{\rm B\kern-.05em{\sc i\kern-.025em b}\kern-.08em
    T\kern-.1667em\lower.7ex\hbox{E}\kern-.125emX}}

\begin{document}
\bstctlcite{IEEEexample:BSTcontrol}

\title{
Recursive Binding for Similarity-Preserving Hypervector Representations of Sequences
\thanks{
The work of DK was supported by the European Union's Horizon 2020 Programme under the MSCA Individual Fellowship Grant (839179). 
DK was also supported in part by AFOSR FA9550-19-1-0241 and Intel's THWAI program.
The work of DAR was supported in part by the National Academy of Sciences of Ukraine (grant 0121U000016), the Ministry of Education and Science of Ukraine (grant no. 0121U000228 and 0122U000818), and the Swedish Foundation for Strategic Research (SSF, grant no. UKR22-0024).
}
}


\author{
\IEEEauthorblockN{Dmitri~A.~Rachkovskij}
\IEEEauthorblockA{\textit{International Research and }\\ 
\textit{Training Center for Information}\\
\textit{Technologies and Systems,} 
Kiev, Ukraine \\
\textit{Lule\aa{} University of Technology,} Lule\aa{}, Sweden\\ 
}
\and
\IEEEauthorblockN{Denis Kleyko}
\IEEEauthorblockA{
\textit{University of California at Berkeley, Redwood }\\
\textit{Center for Theoretical Neuroscience,}  Berkeley, USA \\
 \textit{Research Institutes of Sweden,}\\ 
 \textit{Intelligent Systems Lab,} Kista, Sweden \\
}
}

\maketitle

\begin{abstract}

Hyperdimensional computing (HDC), also known as vector symbolic architectures (VSA), is a computing framework used within artificial intelligence and cognitive computing that operates with distributed vector representations of large fixed dimensionality. A critical step in designing the HDC/VSA solutions is to obtain such representations from the input data. Here, we focus on a wide-spread data type of sequences and propose their transformation to distributed representations that both preserve the similarity of identical sequence elements at nearby positions and are equivariant with respect to the sequence shift. These properties are enabled by forming representations of sequence positions using recursive binding as well as superposition operations. The proposed transformation was experimentally investigated with symbolic strings used for modeling human perception of word similarity. The obtained results are on a par with more sophisticated approaches from the literature. The proposed transformation was designed for the HDC/VSA model known as Fourier Holographic Reduced Representations. However, it can be adapted to some other HDC/VSA models. 

\end{abstract}


\section{Introduction}

Both symbolic and connectionist approaches to Artificial Intelligence have their own advantages. Hyperdimensional computing (HDC~\cite{KanervaHyperdimensional2009}), also known as Vector-Symbolic Architectures (VSA~\cite{GaylerJackendoff2003}) tries to combine the advantages of symbolic structured data representations and connectionist distributed vector representations. There is number of applications at the intersection of electrical engineering, machine learning, and cognitive computing, where HDC/VSA have demonstrated to be a promising approach (see, e.g.,~\cite{RahimiNanoscalable2017, NeubertRobotics2019, RahimiBiosignal2019, SchlegelVSAComparison2020, GeClassificationReview2020, NeubertAggregation2021, hassan2021hyper,KleykoComputingParadigm2021, NeubertPlaceRecognition2021, KleykoSurveyVSA2021Part1, KleykoSurveyVSA2021Part2}). 

In HDC/VSA, distributed vector representations of large fixed dimension, the so-called hypervectors (used with the abbreviation HV interchangeably below), are the main representational unit. The typical requirement is to form HVs in a similarity-preserving fashion. This is important in many settings, e.g., when solving classification problems or performing similarity search. Therefore, design of similarity-preserving transformations of various types of data into HVs has been studied in numerous works, e.g.,~\cite{ KussulRandom1999, KleykoHolographic2017, RachkovskijSimilarityRP2015, RachkovskijAnalogical2004, RachkovskijAnalogy2012, FradyFunctions2021, FradyFunctionsNICE2022, KomerContinuous2019}. 

Sequences are a particularly common type of data. Naturally, there are numerous applications that work with sequential data representation, including but not limited to natural language processing, acoustical and handwriting recognition, bioinformatics, clustering, and information retrieval (e.g.,~\cite{ NavarroGuided2001, YuString2016, KussulDiagnostics1998, BandaragodaTrajectoryTraffic2019, GoltsevCombination2005, RachkovskijIndex2019,DiaoGLVQHD2021} and references therein). These applications rely on measuring similarity of different sequences. The advantage of forming similarity-preserving HVs of sequences is that such representations can possibly be used with a plethora of methods from statistical learning, linear algebra, and computer science that were developed specifically for vectors.

However, most of the known techniques for transforming sequences into HVs are inadequate if the representation should simultaneously preserve similarity of the same elements at nearby positions and be equivariant to the sequence shift. In order to address this limitation, we make a proposal for a transformation satisfying both of the requirements. Our methods are based on the recursive application of a binding operation to represent the order of sequence elements and were developed for the HDC/VSA model of Fourier Holographic Reduced Representations (FHRR)~\cite{PlateHolographic2003}. The proposed methods can be adapted for other HDC/VSA models that use multiplicative and recursive binding operation (see~\cite{SchlegelVSAComparison2020, KleykoSurveyVSA2021Part1}).

The main contributions of this paper are as follows: 
\noindent
\begin{itemize}
    \item Recursive role binding-based hypervector representation of sequences that is shift-equivariant and preserves the similarity of sequences with the same elements at nearby positions;
    \item Similarity measures for sequences based on hypervectors;
    \item  Experimental study of the proposed hypervector representations of sequences and similarity measures in two benchmarks from the area of modeling human perception of word similarity~\cite{HannaganHolographic2011, HannaganProtein2012, DavisSpatial2010}.
\end{itemize}

\section{Background}
\label{sect:background}

\subsection{Hyperdimensional Computing/Vector Symbolic Architectures}
\label{sect:supp:vsa}

Here, we provide a brief summary of HDC/VSA (for a comprehensive survey, please advise~\cite{KleykoSurveyVSA2021Part1, KleykoSurveyVSA2021Part2}).
There are various HDC/VSA models. One of the main differences between these models is the format of hypervector components. Examples are real numbers from the Gaussian distribution (the HRR~\cite{PlateHolographic2003} and the MBAT~\cite{GallantRepresenting2013} models), complex numbers (the FHRR model~\cite{PlateHolographic2003}), or binary values from {0,1} (the BSC~\cite{KanervaOrdered1996}, SBDR~\cite{ RachkovskijBinding2001, KleykoSDR2016}, and SBC~\cite{LaihoSparse2015, FradySDR2020} models).

Random i.i.d. HVs are called atomic as they are used as the elementary building block of representation. 
This paper uses the FHRR model. In FHRR, dense complex HVs are used. In particular, in the atomic HVs, each component has the unit magnitude and phase chosen at random.
Data HVs are formed from the atomic HVs of the data elements, usually without changing the HV dimension. For example, for elements-symbols, their HVs are random atomic of high dimension $D$, commonly $D > 1000$.

The atomic HVs are combined using some combination of the key HDC/VSA operations: superposition and binding. The superposition a.k.a bundling operation (denoted as $+$) is implemented via component-wise addition. It is commonly used to represent a set (e.g., a set of symbols). The superposition operation does not preserve information about the grouping or order of the elements, but the result of superposition is similar to the HVs that are superimposed.

The multiplicative binding operation (denoted as $\odot$) is implemented via component-wise multiplication. The binding operation is important for representing order or grouping information. For example, when representing a sequence, the HV of its symbol at some position (``filler'') is bound to the position's HV (``role''). Thus, the result of the binding operation is an HV containing information about role and filler HVs used for the binding operation. However, in FHRR the result of the binding operation will be dissimilar to both role and filler HVs. Also, for the bound HVs to be similar, their role HVs should be similar to each other, and the filler HVs should be similar as well. 
The binding operation distributes over the superposition operation. It is worth pointing out that binding can also be implemented by permutations, but we do not exploit permutations in the transformation proposed in this work.

The binding and superposition operations can be used to represent a sequence in HV, e.g., as follows. Given a sequence instance such as a symbolic string, one first forms the bindings of its symbols' HVs with the HVs of their corresponding positions. These bindings are then superimposed into a single HV representing the whole sequence. We discuss some of the varieties of the schemes for representing sequences in section~\ref{sec:related}.
Finally, the similarity between HVs should be measured using some quantitative measure. The similarity measures used in this paper are introduced below.

\subsection{Equivariance of hypervector representations}

Let $x$ be an object (input), $F$ be a function performing a representation, $F(x) $ be the result of $x$ representation. $F$ is equivariant with respect to transformations $T$, $S$ if~\cite{CohenGroupCNN2016}: $F(S(x)) = T(F(x)) $. Transformations $T$ and $S$ can be different.
The advantages of equivariant representations are considered, e.g., in~\cite{CohenGroupCNN2016} and~\cite{RachkovskijEquivariant2021}. For example, the system need not to represent each input datum if it is transformed in some way and its transformed representation can be obtained from the available one. In the context of brain studies, this can be considered as mental transformation or mental imagery~\cite{pearson2015mental, christophel2015parietal}.

We consider hypervector representations of sequences. Sequences can be transformed using shift operations. Let us represent the sequence $x$ as an HV by applying some function (algorithm) $F(x) $. Then shift $x$ to another position, denote this transformation by $S(x) $. The HV of the shifted sequence is obtained as $F(S(x)) $. The representation function $F$, that is equivariant with respect to $S(x) $, must ensure $F(S(x)) = T(F(x)) $, where $T$ is some transformation of the HV $F(x) $. In other words, the HV of the shifted sequence can be obtained not only by transforming this sequence into an HV, but also by some transformation of the HV of the unshifted sequence. 

In this paper, we consider sequences of symbols from a finite alphabet. Denote as $x_i$ the symbol (sequence element) $x$ at position $i$. For instance, $a_0$ denotes $a$ at the beginning of the string (at the initial position), $a_{-1}$ is the same symbol shifted one position to the left, $b_3$ is the symbol $b$ shifted $3$ positions to the right, and so on. If a symbol is specified without an index, it is assumed to be at the initial position: $x \equiv x_0$.

We denote by $x_iy_j ... z_k$ the sequence of symbols $x,y, ..., z$ that are located at positions $i, j, ..., k$, respectively (e.g., $b_3c_1a_4a_{-3}$). A symbolic string (symbols at consecutive positions) is denoted as $x_i y_{i+1} ... z_{i+k} \equiv xy...z_i$, e.g., $c_1b_2c_3a_4 \equiv cbca_1 \equiv (cbca)_1$. A string without an index is at its initial position, e.g., $cbca \equiv cbca_0 \equiv c_0b_1c_2a_3$.

HVs corresponding to symbols/sequences will be denoted by the corresponding bold letters. For example, $F(a_0) = \mathbf{a}_0$; $F(cbca_0) = \mathbf{cbca}_0$; $F(cbca_s) = \mathbf{cbca}_s$. We denote the shift of symbols by $s$ positions by $S_s$: $S_1(a_0) = a_1$, $S_{-1}(abc) = S_{-1}(abc_0) = abc_{-1} \equiv a_{-1}b_0c_1$. Let $T_s$ denote the HV transform corresponding to $S_s$. To ensure equivariance, the following must be true: $F(S_s(x)) = T_s(F(x)) = \mathbf{x}_s$ ($x$ is a symbol or sequence). For instance, $F(S_1(a_0)) = \mathbf{a}_1 = T_1(F(a_0)), F(S_2(abc_0)) = T_2(F(abc_0)) = \mathbf{abc}_2$, $F(S_{-2}(abc_4)) = T_{-2}(F(abc_4)) = \mathbf{abc}_2$, etc. 
Some of the existing approaches to the hypervector representations of sequences that are shift-equivariant are considered in section~\ref{sec:related}.

\section{The proposed transformation}

To preserve both the equivariance of HV representations of sequences with respect to the shift and the similarity of the sequence HVs having the same symbols at nearby positions, we propose to form the HVs of positions as compositional HVs of a specific structure, using the application of recursive binding and superposition. 

\subsection{Hypervector representation of symbols}
\label{sec:trans:symb}

Let us generate a random (``atomic'') HV $\mathbf{pos}$ to be used for the representations of positions. 
To represent the symbol $a$, we will form its HV $\mathbf{a}$ as follows. 
First, we generate another atomic HV $\mathbf{e}_{a 0} \equiv \mathbf{e}_a$. 
Let us form other atomic HVs as:
\noindent
\begin{equation}
\mathbf{e}_{a i} = \mathbf{pos} \odot \mathbf{e}_{a i-1} =  \mathbf{pos}^i \odot \mathbf{e}_a
\label{eq:hv:symb} 
\end{equation}
\noindent
Note that for $i < 0$, $\mathbf{pos}^i = \mathbf{\overline{pos}}^{|i|}$ (complex conjugate). 
Next, we obtain the HV of the symbol $a$ at position $i$ (that is, $\mathbf{a}_i = F(a_i)$) for a given value of the “similarity radius” $R\geq 1$:
\noindent
\begin{equation}
\begin{split}
&\mathbf{a}_i =  \mathbf{e}_{a i} + \mathbf{e}_{a i+1} + \cdots + \mathbf{e}_{a i+R-1} = \\
& \mathbf{pos}^{i} \odot \mathbf{e}_a + \mathbf{pos}^{i+1} \odot \mathbf{e}_a + \cdots + \mathbf{pos}^{i+R-1} \odot \mathbf{e}_a =  \\
&   \mathbf{e}_a \odot( \mathbf{pos}^{i} + \mathbf{pos}^{i+1} + \cdots + \mathbf{pos}^{i+R-1}).
\end{split}
\label{eq:hv:symb:pos} 
\end{equation}
\noindent

\begin{figure}[t]
    \centering
    \includegraphics[width=1.0\columnwidth]{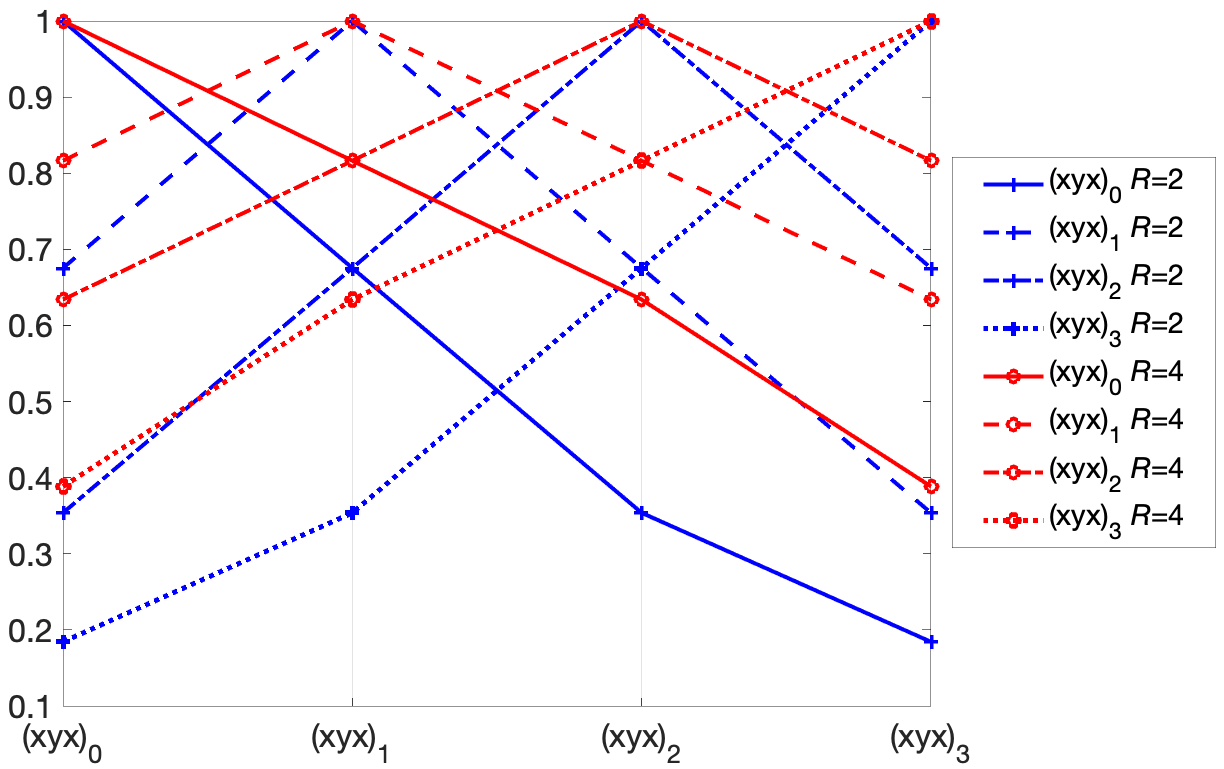}
    \caption{
The HV similarity $\text{sim}_{\text{HV,cos}}$ values of the string $xyx$ with itself at various positions, $D=10000$.
    }
    \label{fig:HV:str:sim}
\end{figure}

\subsubsection{Equivariance}
\label{sec:trans:symb:equiv}

For such an HV representation, the equivariance with respect to the symbol shift is satisfied if an appropriate recursive binding is used as the HV transformation $T$. Indeed,
\noindent
\begin{equation}
\begin{split}
& T_j(F(a_i))  = \mathbf{pos}^j \odot \mathbf{a}_i =  \\
& \mathbf{e}_{a} \odot \mathbf{pos}^j \odot ( \mathbf{pos}^{i} + \mathbf{pos}^{i+1} + \cdots +  \mathbf{pos}^{i+R-1}  )=\\
& \mathbf{e}_{a} \odot ( \mathbf{pos}^{i+j} + \mathbf{pos}^{i+j+1} + \cdots +  \mathbf{pos}^{i+j+R-1}  )=\\
& \mathbf{e}_{a i+j} + \mathbf{e}_{a i+j+1} + \cdots + \mathbf{e}_{a i+j+R-1} =\\
&\mathbf{a}_{i+j} = F(a_{i+j}) =F (S_j(a_i)). 
\end{split}
\label{eq:hv:equivariance} 
\end{equation}
\noindent

\subsubsection{Similarity}
Let us consider HVs: $\mathbf{a}_i =  \mathbf{e}_{a i} + \mathbf{e}_{a i+1} + \cdots + \mathbf{e}_{a i+R-1}$ and $\mathbf{a}_{i+j} =  \mathbf{e}_{a i+j} + \mathbf{e}_{a i+j+1} + \cdots + \mathbf{e}_{a i+j+R-1}$.
For $|j| < R$, $\mathbf{a}_{i}$ and $\mathbf{a}_{i+j}$ have $R - |j|$ coinciding atomic HVs. 
For example, for $j > 0$ these are atomic HVs with indices from $i + j$ to $i + R - 1$ (the last atomic HVs from $\mathbf{a}_{i}$ and the first atomic HVs from $\mathbf{a}_{i+j}$; for $j < 0$, the opposite is true). 
For $|j| \geq R$, $\mathbf{a}_{i}$ and $\mathbf{a}_{i+j}$ have no coinciding atomic HVs. This is reflected in the value of the similarity measure between $\mathbf{a}_i$ and $\mathbf{a}_{i+j}$ when it is calculated based on the dot product~\cite{FradyCapacity2018}.

\subsection{Hypervector representation and similarity of sequences}

HVs of various symbols at their positions are formed by the method of section~\ref{sec:trans:symb} from their randomly generated atomic HVs, using the same $\mathbf{pos}$. A symbol sequence HV is formed from the symbol HVs using the binding and superposition operations: 
$\mathbf{x}_i\mathbf{y}_j ... \mathbf{z}_k = F(x_iy_j ... z_k) = \mathbf{pos}^{i} \odot \mathbf{x}_0 + ... + \mathbf{pos}^{k} \odot \mathbf{z}_0$.
The properties of equivariance and preservation of similarity for HVs of symbol sequences can be obtained in the same manner as in section~\ref{sec:trans:symb}. This is achieved due to the distributive property of multiplicative binding over the superposition operation.

\begin{figure}[t]
    \centering
    \includegraphics[width=1.0\columnwidth]{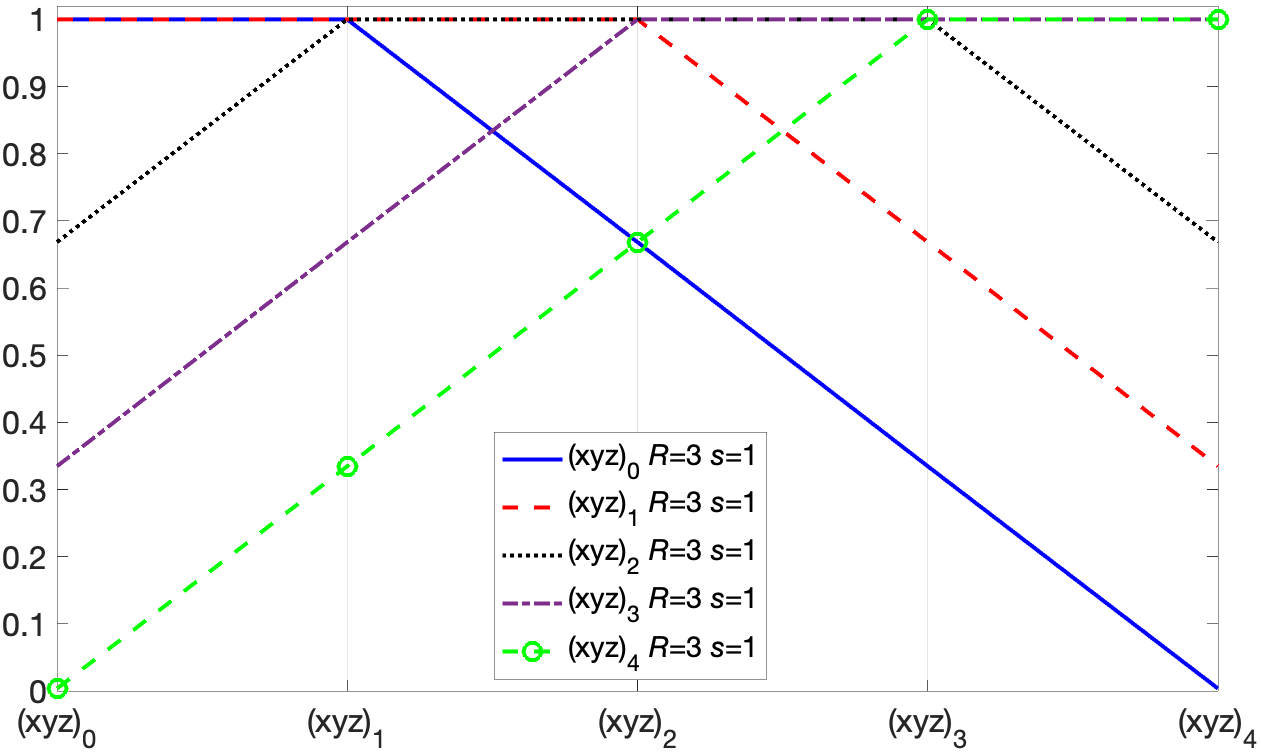}
    \caption{
The values of $\text{sim}_{\text{HV,cos}}$ between the HV representation of the string $xyz$ at different positions, taking into account the shift; $D = 10000$. 
    }
    \label{fig:HV:str:sim:shift}
\end{figure}

\subsubsection{Hypervector similarity of strings without shift}

We calculate hypervector similarity using similarity measures of complex-valued vectors. Denote $\mathbf{x} = F(x)$ the HV of $x$ obtained by the method proposed above for some specified $R$. The  similarities can be obtained with, e.g., the following measures. 
\noindent
The cosine similarity: 
\noindent
\begin{equation}
\text{sim}_{\text{HV,cos}} = \Re( \langle \mathbf{x}, \mathbf{y}  \rangle) / \sqrt{\langle \mathbf{x}, \mathbf{x}  \rangle \langle \mathbf{y}, \mathbf{y}  \rangle},
\label{eq:hv:sim:cos} 
\end{equation}
\noindent
where $\langle \cdot, \cdot  \rangle$ is the complex dot product and $\Re()$ denotes its real part.  
Jaccard: 
\noindent
\begin{equation}
\text{sim}_{\text{HV,Jac}} = \Re( \langle \mathbf{x}, \mathbf{y}  \rangle) / 
({\langle \mathbf{x}, \mathbf{x}\rangle } +
{ \langle \mathbf{y}, \mathbf{y}  \rangle} -  \Re( \langle \mathbf{x}, \mathbf{y}  \rangle) ).
\label{eq:hv:sim:jac} 
\end{equation}
\noindent
Simpson:
\noindent
\begin{equation}
\text{sim}_{\text{HV,Simp}} = \Re( \langle \mathbf{x}, \mathbf{y}  \rangle) / 
\min ({\langle \mathbf{x}, \mathbf{x}\rangle },
{ \langle \mathbf{y}, \mathbf{y}  \rangle} ).
\label{eq:hv:sim:sim} 
\end{equation}
\noindent
Examples of HV similarity characteristics for a string at different positions are shown in Fig.~\ref{fig:HV:str:sim}.

\subsubsection{Hypervector similarity of strings with the shift}

Strings might have identical substrings outside $R$. For instance, for $dddabc_0$ and $abc_0$, the value of $\text{sim}_{\text{HV}}$ is close to zero for $R\geq 3$. However, if $abc_0$ is shifted to $abc_3$, the string $abc_3$ will match the substring of $dddabc_0$. Let us take into account such cases by calculating the similarity as the maximum value of $\text{sim}_{\text{HV}}$ for various shifts of one of the sequences: 
\noindent
\begin{equation}
\begin{split}
& \text{sim}_{\text{HV},s} (\mathbf{x},\mathbf{y}) = \max_{s} \text{sim}_{\text{HV}} (S_s(x),y) = \\
& \max_{s} \text{sim}_{\text{HV}} (F(S_s(x)),F(y)) = \\
& \max_{s} \text{sim}_{\text{HV}} (T_s(F(x)),F(y)) = \max_{s} \text{sim}_{\text{HV}} (T_s(\mathbf{x}),\mathbf{y}).
\end{split}
\label{eq:sim:shift} 
\end{equation}
\noindent
\parbox{\columnwidth}{Unless stated otherwise, we assume that the numeric value $s$ specifies the set of shifts from $-s$ to $s$ in steps of $1$. For example, if $s = 1$ then $\text{sim}_{\text{HV},s} (x,y)$ is the max value of 
\parfillskip=0pt}

\begin{landscape}

\setlength\tabcolsep{1.5pt} 
\renewcommand{\arraystretch}{1.236}
\begin{table}[t]
    \centering
    \caption{
The similarities of different word pairs and satisfaction of similarity requirements (columns ``?'') by various models.
In~\cite{CohenOrthogonality2013}, $\text{Ed} = 1-\text{dist}_{\text{Lev}}(x,y)/(|x|+|y|)$ and ``C'' denotes FHRR with  similarity-preserving roles. The columns without references contain our results reported for the cosine similarity. If $s$ is absent, it means $s=0$. Please also see the main text for other abbreviations and details. Columns in bold show the model from~\cite{CoxHolographic2011} and our model for which all similarity requirements are satisfied.
    }
    \label{tab:first}
    \begin{tabular}{| c| c | c |c |c |c |c |c| c | c |c |c |c |c |c| c | c |c |c |c |c |c| c | c |c |c |c |c |c| c | c |c |c |c |c |c |c |}
        \hline
        
    \multicolumn{5}{|c|}{\cite{HannaganHolographic2011}}& \multicolumn{8}{c|}{BSC \cite{HannaganHolographic2011}} & \multicolumn{2}{c|}{\cite{CohenOrthogonality2013}} & \multicolumn{2}{c|}{\cite{CoxHolographic2011}} & \multicolumn{2}{c|}{nodb} &  \multicolumn{2}{c|}{db} & \multicolumn{2}{c|}{HV nodb} & \multicolumn{2}{c|}{HV db}  & \multicolumn{6}{c|}{HV $s=2$ no db} & \multicolumn{6}{c|}{HV $s=2$ db}   
    \\ \hline
    
    Constraint & Prime & Target & Cond & Criteria & Slot & ? & COB & ? &	UOB & ? &	LCD &	? & ?C &	?Ed &	TR &	? &	-Lev &	? &	-Lev & ? & $R=1 $ &	? &	$R=1$ &	? &	$R=1$ &	? &	$R=2$ &	? &	$R=3$ &	? &	$R=1$ &	? &	$R=2$ &	? &	$R=3$ &	?  \\  \hline

    \multirow{8}{*}{Stability}  & 12345 &	12345 &	(1) &	$>0.95$ &	1.00 &	Y &	1.00 &	Y &	1.00 &	Y &	1.00 &	Y &	Y &	Y &	\textbf{1.0} &	\textbf{Y} &	0 &	Y &	0 &	Y &	1.00 &	Y &	1.00 &	Y &	1.00 &	Y &	1.00 &	Y &	1.00 &	Y &	1.00 &	Y &	1.00 &	Y &	\textbf{1.00} &	\textbf{Y}
   \\ \cline{2-37} 
    & 1245 &	12345 &	(2) &	$<(1)$ &	0.28 &	Y &	0.6 &	Y &	0.62 &	Y &	0.53 &	Y &	Y &	Y &	\textbf{0.8} &	\textbf{Y} &	-1 &	Y &	-1 &	Y &	0.45 &	Y &	0.62 &	Y &	0.45 &	Y &	0.67 &	Y &	0.75 &	Y &	0.62 &	Y &	0.77 &	Y &	\textbf{0.84} &	\textbf{Y}
    \\ \cline{2-37} 
     &	123345 &	12345 &	(3) &	$<(1)$ &	0.36 &	Y &	0.7 &	Y &	0.79 &	Y &	0.45 &	Y &	Y &	Y &	\textbf{0.9} &	\textbf{Y} &	-1 &	Y &	-1 &	Y &	0.55 &	Y &	0.67 &	Y &	0.55 &	Y &	0.76 &	Y &	0.83 &	Y &	0.67 &	Y &	0.80 &	Y &	\textbf{0.87} &	\textbf{Y} \\  \cline{2-37} 
    
     &	123d45 &	12345 &	(4) &	$<(1)$ &	0.36 &	Y &	0.7 &	Y &	0.73 &	Y &	0.51 &	Y &	Y &	Y &	\textbf{0.8} &	\textbf{Y} &	-1 &	Y &	-1 &	Y &	0.55 &	Y &	0.67 &	Y &	0.55 &	Y &	0.73 &	Y &	0.79 &	Y &	0.67 &	Y &	0.79 &	Y &	\textbf{0.85} &	\textbf{Y} \\  \cline{2-37} 
     &	12dd5 &	12345 &	(5) &	$<(1)$ &	0.43 &	Y &	0.4 &	Y &	0.45 &	Y &	0.37 &	Y &	Y &	Y &	\textbf{0.4} &	\textbf{Y} &	-2 &	Y &	-2 &	Y &	0.60 &	Y &	0.71 &	Y &	0.60 &	Y &	0.55 &	Y &	0.53 &	Y &	0.71 &	Y &	0.74 &	Y &	\textbf{0.74}&	\textbf{Y} \\  \cline{2-37} 
    
     &	1d345 &	12345 &	(6) &	$<(1)$ &	0.76 &	Y &	0.8 &	Y &	0.79 &	Y &	0.78 &	Y &	Y &	Y &	\textbf{0.6} &	\textbf{Y} &	-1 &	Y &	-1 &	Y &	0.80 &	Y &	0.86 &	Y &	0.80 &	Y &	0.80 &	Y &	0.80 &	Y &	0.86 &	Y &	0.89 &	Y &	\textbf{0.90} &	\textbf{Y} \\  \cline{2-37} 
    
     &	12d456 &	123456 &	(7) &	$<(1)$ &	0.83 &	Y &	0.8 &	Y &	0.85 &	Y &	0.80 &	Y &	Y &	Y &	\textbf{0.7} &	\textbf{Y} &	-1 &	Y &	-1 &	Y &	0.83 &	Y &	0.87 &	Y &	0.83 &	Y &	0.83 &	Y &	0.83 &	Y &	0.87 &	Y &	0.90 &	Y &	\textbf{0.91} &	\textbf{Y} \\  \cline{2-37} 
     &	12d4d6 &	123456 &	(8) &	$<(7)$ &	0.67 &	Y &	0.8 &	Y &	0.75 &	Y &	0.70 &	Y &	Y &	Y &	\textbf{0.4} &	\textbf{Y} &	-2 &	Y &	-2 &	Y &	0.67 &	Y &	0.75 &	Y &	0.67 &	Y &	0.67 &	Y &	0.63 &	Y &	0.75 &	Y &	0.80 &	Y &	\textbf{0.79} &	\textbf{Y}  \\ \hline
    \multirow{3}{*}{Edge}   &	d2345 &	12345 &	(9) &	$<(10)$ &	0.62 &	N &	0.7 &	N &	0.49 &	Y &	0.56 &	Y &	N &	N &	\textbf{0.5} &	\textbf{Y} &	-1 &	N &	-2 &	Y &	0.80 &	N &	0.71 &	Y &	0.80 &	N &	0.80 &	N &	0.80 &	N &	0.71 &	Y &	0.67 &	Y &	\textbf{0.65} &	\textbf{Y}  \\ \cline{2-37} 
     &	12d45 &	12345 &	(10) &	$<(1)$ &	0.62 &	Y &	0.6 &	Y &	0.62 &	Y &	0.63 &	Y &	N &	N &	\textbf{0.7} &	\textbf{Y} &	-1 &	Y &	-1 &	Y &	0.80 &	Y &	0.86 &	Y &	0.80 &	Y &	0.80 &	Y &	0.80 &	Y &	0.86 &	Y &	0.89 &	Y &	\textbf{0.90} &	\textbf{Y}  \\ \cline{2-37} 
     &	1234d &	12345 &	(11) &	$<(10)$ &	0.62 &	N &	0.7 &	N &	0.53 &	Y &	0.56 &	Y &	N &	N &	\textbf{0.5} &	\textbf{Y} &	-1 &	N &	-2 &	Y &	0.80 &	N &	0.72 &	Y &	0.80 &	N &	0.80 &	N &	0.80 &	N &	0.72 &	Y &	0.67 &	Y &	\textbf{0.65} &	\textbf{Y}  \\ \hline
    TL local &	12435 &	12345 &	(12) &	$>(5)$ &	0.62 &	Y &	0.8 &	Y &	0.82 &	Y &	0.79 &	Y &	Y &	Y &	\textbf{0.7} &	\textbf{Y} &	-2 &	N &	-2 &	N &	0.60 &	N &	0.71 &	N &	0.60 &	N &	0.80 &	Y &	0.87 &	Y &	0.71 &	N &	0.89 &	Y &	\textbf{0.93} &	\textbf{Y}  \\ \hline
    TL global &	21436587 &	12345678 &	(13) &	$Min$ &	0.00 &	Y &	0.7 &	N &	0.79 &	N &	0.62 &	N &	N &	Y &	\textbf{0.3} &	\textbf{Y} &	-5 &	Y &	-7 &	Y &	-0.0 &	Y &	-0.0 &	Y &	0.51 &	N &	0.50 &	Y &	0.67 &	N &	0.41 &	Y &	0.42 &	Y &	\textbf{0.53} &	\textbf{Y}  \\ \hline
    TL distant &	125436 &	123456 &	(14) &	$<(7)>(8)$ &	0.84 &	N &	0.8 &	N &	0.92 &	N &	0.87 &	N &	N &	N &	\textbf{0.7} &	\textbf{Y} &	-2 &	N &	-2 &	N &	0.67 &	N &	0.75 &	N &	0.67 &	N &	0.67 &	N &	0.78 &	Y &	0.75 &	N &	0.80 &	N &	\textbf{0.88} &	\textbf{Y}  \\ \hline
    TL compound &	13d45 &	12345 &	(15) &	$<(6)$ &	0.67 &	Y &	0.8 &	N &	0.79 &	N &	0.78 &	N &	Y &	N &	\textbf{0.5} &	\textbf{Y} &	-2 &	Y &	-2 &	Y &	0.60 &	Y &	0.71 &	Y &	0.60 &	Y &	0.70 &	Y &	0.73 &	Y &	0.71 &	Y &	0.83 &	Y &	\textbf{0.86} &	\textbf{Y}  \\ \hline
    \multirow{4}{*}{RP distinct}  &	12345 &	1234567 &	(16) &	$>Min$ &	0.68 &	Y &	0.6 &	N &	0.52 &	N &	0.58 &	N &	Y &	Y &	\textbf{0.6} &	\textbf{Y} &	-2 &	Y &	-3 &	Y &	0.84 &	Y &	0.76 &	Y &	0.84 &	Y &	0.85 &	Y &	0.85 &	Y &	0.76 &	Y &	0.75 &	Y &	\textbf{0.75} &	\textbf{Y}  \\ \cline{2-37} 
     &	34567 &	1234567 &	(17) &	$>Min$ &	0.00 &	Y &	0.6 &	N &	0.51 &	N &	0.58 &	N &	Y &	Y &	\textbf{0.6} &	\textbf{Y} &	-2 &	Y &	-3 &	Y &	0.01 &	? &	0.01 &	? &	0.85 &	Y &	0.85 &	Y &	0.85 &	Y &	0.75 &	Y &	0.75 &	Y &	\textbf{0.75} &	\textbf{Y}  \\ \cline{2-37} 
    &	13457 &	1234567 &	(18) &	$>Min$ &	0.12 &	Y &	0.6 &	N &	0.61 &	N &	0.49 &	N &	Y &	Y &	\textbf{0.6} &	\textbf{Y} &	-2 &	Y &	-2 &	Y &	0.17 &	Y &	0.25 &	Y &	0.51 &	N &	0.68 &	Y &	0.73 &	Y &	0.63 &	Y &	0.70 &	Y &	\textbf{0.79} &	\textbf{Y}  \\ \cline{2-37} 
    &	123256 &	1232456 &	(19) &	$>Min$ &	0.70 &	Y &	0.8 &	Y &	0.89 &	Y &	0.82 &	Y &	Y &	Y &	\textbf{0.9} &	\textbf{Y} &	-1 &	Y &	-1 &	Y &	0.62 &	Y &	0.71 &	Y &	0.62 &	Y &	0.77 &	Y &	0.84 &	Y &	0.71 &	Y &	0.81 &	Y &	\textbf{0.87} &	\textbf{Y}  \\ \hline
    RP repeated &	123456 &	1232456 &	(20) &	$=(19)$ &	0.65 &	N &	0.8 &	Y &	0.89 &	Y &	0.78 &	N &	Y &	Y &	\textbf{0.9} &	\textbf{Y} &	-1 &	Y &	-1 &	Y &	0.47 &	N &	0.59 &	N &	0.47 &	N &	0.77 &	Y &	0.84 &	Y &	0.59 &	N &	0.81 &	Y &	\textbf{0.87} &	\textbf{Y}  \\ \hline

    \end{tabular}
\end{table}

\begin{multicols}{3}

\noindent
$\text{sim}_{\text{HV},s} (S_s(x),y)$ obtained with the shifts $\{-1,0,1\}$ of the sequence $x$. Shift equivariance permits obtaining the HVs of shifted sequences by binding with $\mathbf{pos}^s$ the sequence HV obtained at the initial position. Examples of the string HV similarity (\ref{eq:sim:shift}) characteristics are shown in Fig.~\ref{fig:HV:str:sim:shift}.

\section{Empirical evaluation}

Experimental evaluations of the proposed approach were carried out in the tasks that model the identification of visual images of words by humans. In particular, we present the results of experiments on the similarity of words using their HV representation. The results are compared to those obtained by psycholinguists for human subjects~\cite{DavisSpatial2010,CoxHolographic2011}. This task was chosen as its results depend significantly on both parameters $R$ and $s$. It is important to note that the main intention of the experiments was to demonstrate the feasibility of the proposed transformation.  

The experiments investigated the priming (effect of precedence) for visual (printed) words in humans: how different priming letter strings influenced the speed of perception (identification) of targets, e.g., deciding whether a target is a familiar word. Due to the short-term exposure of primes (50 ms), the processing of information about them by the subjects did not include conscious processes. 

Certain types of primes facilitate processing of the target string, i.e., decrease the delay (compared to a neutral prime) before producing a correct response. The relative amount of facilitation by different types of primes can be considered as a measure of the relative similarity of the prime and the target, regardless of which representation is used by the visual word recognition system in humans. Thus, visual word representation models should strive to ensure that their representations reflect the similarity patterns obtained in priming experiments with humans~\cite{HannaganHolographic2011}.

\subsection{Modeling restrictions on the perception of word similarity}

The properties of word similarity obtained by psycholinguists in experiments with human subjects have been summarized in~\cite{HannaganHolographic2011}. 
These similarity properties were categorized into four types of restrictions (Table~\ref{tab:first} shows how these restrictions were satisfied by various approaches):

\end{multicols}

\end{landscape}

\noindent
\begin{itemize}
    \item Stability: the similarity of a string to any other string is less than to itself;
    \item Edge effect: the greater importance of the outer letters coincidence compared to the inner ones;
    \item Transposed letter effect: transposing letters reduces similarity less than replacing them with other letters;
    \item Relative position effect: breaking the absolute letter order while keeping the relative one still gives effective priming.
\end{itemize}

\noindent
In~\cite{HannaganHolographic2011}, the authors have provided examples of string pairs for each similarity type and the relations between the similarity values for different pairs. We used these data in our experiments (see columns Prime, Target, Criteria in Table~\ref{tab:first}). Table~\ref{tab:first} presents the results obtained by previous studies as well as the results for the proposed transformation.

Our results were obtained for HVs with $D=10000$, as the mean HV similarity values of $50$ random realizations. To reflect the edge effect, we used the ``db'' option: the HVs were formed in a special way equivalent to the HV representation of strings with doubled first and last letters. For $s > 0$, the hypervectors corresponding to the (left and right) shifted strings were obtained by the proposed transformation of the string HV at $s = 0$. We also show the results obtained with the negative Levenshtein string distance ($-\text{dist}_{\text{Lev}}$). The best fit to the human constraints was for $R = \{2,3\}$, $s = 2$ (Table~\ref{tab:first}). All constraints were satisfied for $R = 3$, $s = 2$. 

\begin{figure}[t]
    \centering
    \includegraphics[width=1.0\columnwidth]{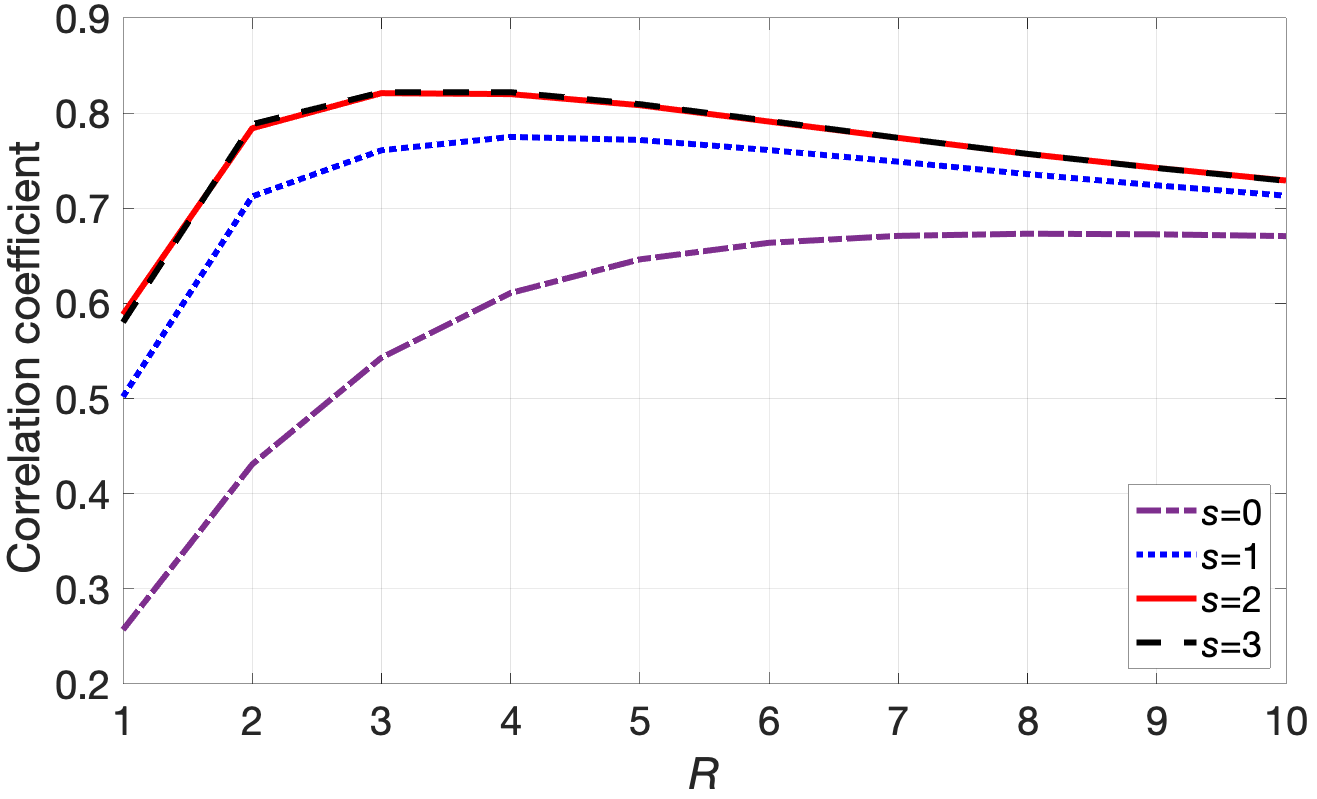}
    \caption{
Mean values of the Pearson correlation coefficient between the HV similarity values (cosine) and the times of the forward priming vs $R$.
    }
    \label{fig:human:corr}
\end{figure}

For comparison, the results of the following models are shown:
\begin{enumerate}
    \item In~\cite{HannaganHolographic2011}, the HV representations of the BSC model~\cite{KanervaOrdered1996} were used (dense binary HVs, binding via component-wise XOR, superposition by addition with threshold binarization, i.e., the majority rule) and their similarity based on $\text{dist}_{\text{Ham}}$  of binary vectors: $1 - 2 \text{dist}_{\text{Ham}}/D$. The following sequence representations were used: 
    \begin{itemize}
        \item Slot: superposition of HVs obtained by binding HVs of each letter and a random HV of its position in the string;

        \item Constrained open-bigrams (COB): all subsequences of two letters with a position difference of up to 3, each subsequence $ab$ is represented as $\mathbf{a} \odot \mathbf{left} + \mathbf{b} \odot \mathbf{right}$, the HVs of all subsequences are superimposed;
        
        \item  Unconstrained open-bigrams (UOB): all subsequences of two letters;
        \item  Local combination detector (LCD): a combination of Slot and COB;
        \item  Seriol: Advanced COB;
        \item  Spatial Coding: see~\cite{DavisSpatial2010}. 
    \end{itemize}
    
     \item~\cite{CohenOrthogonality2013} used the HVs of the BSC model as well as complex-valued HVs of FHRR and real-valued HVs of HRR~\cite{PlateHolographic2003} (in HRR, the binding is the cyclic convolution, the superposition is the normalized addition). 
     String HVs were formed as a superposition of HVs obtained by binding symbol HVs with HVs of their positions. 
     In contrast to~\cite{HannaganHolographic2011}, position HVs were correlated, and their similarity decreased linearly along the length of a particular word. 
     As a measure of similarity for binary HVs, $1 - 2 \text{dist}_{\text{Ham}}/D$ was used, and $\text{sim}_{\text{HV,cos}}$ for complex and real-valued HVs. In Table~\ref{tab:first}, we report their best results obtained with FHRR.

    \item~\cite{CoxHolographic2011} proposed a terminal-relative (TR) string representation scheme. It used not only the representation of letters without position, but also bi-grams, as well as the representation of the positions of letters and bi-grams relative to the terminal letters of the word. The HV representation of the TR scheme was implemented in the HRR model, $\text{sim}_{\text{HV,cos}}$ was used as a similarity measure. To represent the bi-gram, when binding two letter HVs, different permutations were used for the left and right letter HVs. To represent relative to terminal letters, the terminal letter HV was also bound as the left or the right one. In this case, to represent non-adjacent letters, an additional ``space'' character could be introduced. Both options meet all conditions from~\cite{HannaganHolographic2011}.

 \end{enumerate}

\begin{figure}[t]
    \centering
    \includegraphics[width=0.63\columnwidth]{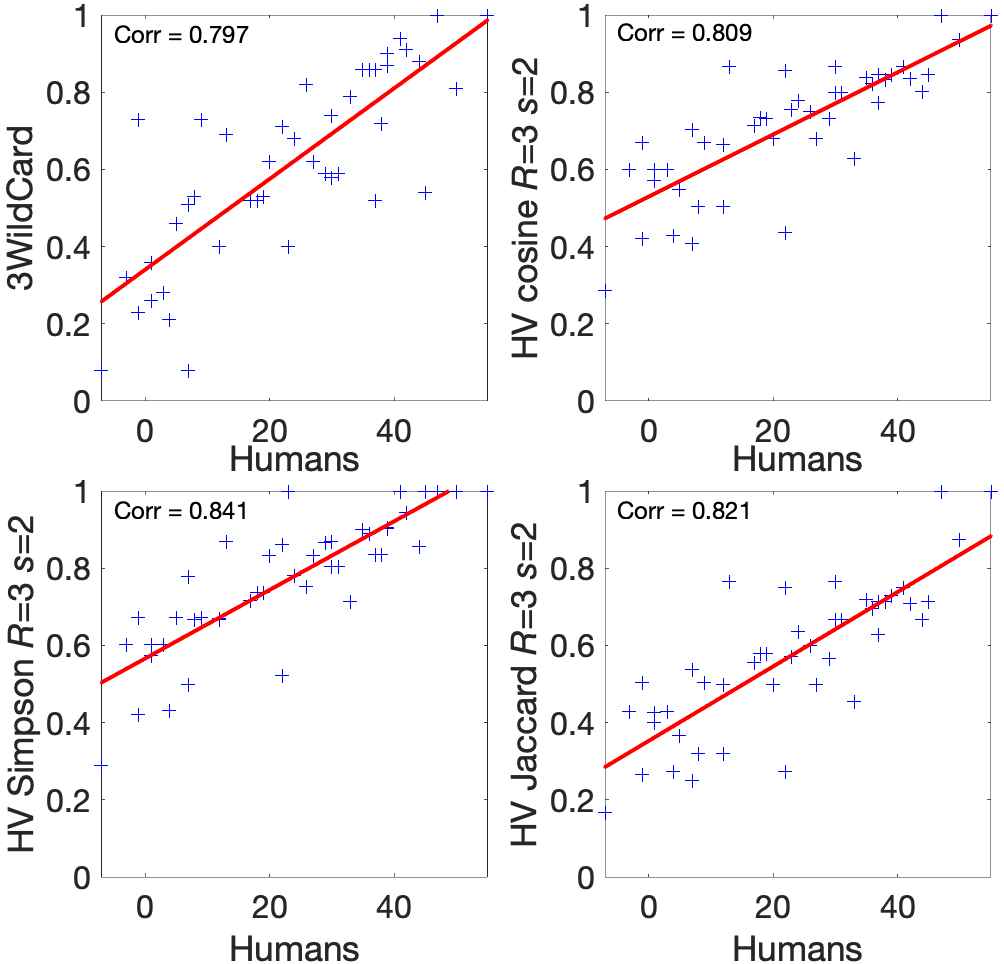}
    \caption{
Scatterplots showing human priming data ($x$-axis, in ms) and similarity values ($y$-axis) obtained for various measures of string similarity. 
    }
    \label{fig:human:data}
\end{figure}

\begin{table}[t]
    \centering
    \caption{
    The Pearson correlation coefficient Corr between the data from the priming experiments with human subjects and the data obtained by various measures of similarity; calculated using the test~\cite{HannaganProtein2012} with $45$ string pairs; sim type Jac stands for (\ref{eq:hv:sim:jac}), whereas Simp stands for (\ref{eq:hv:sim:sim}). 
    }
    \label{tab:HannaganProtein2012}
    \begin{tabular}{| c| c | c |c |c |c |c |}
        \hline
         Method & db & $R$ & $s$ & sim type & Corr & std  \\  \hline
         Spatial coding \cite{HannaganProtein2012} & \cellcolor{gray!25}  & \cellcolor{gray!25} & \cellcolor{gray!25} &  \cellcolor{gray!25} &  $0.73$ & \cellcolor{gray!25} \\  \hline
         GvH  UOB \cite{HannaganProtein2012} &  \cellcolor{gray!25} & \cellcolor{gray!25} & \cellcolor{gray!25} & \cellcolor{gray!25} &  $0.67$ & \cellcolor{gray!25} \\  \hline         
         Kernel UOB \cite{HannaganProtein2012} & \cellcolor{gray!25} & \cellcolor{gray!25} & \cellcolor{gray!25} & \cellcolor{gray!25}  &  $0.75$ & \cellcolor{gray!25} \\  \hline         
         3-Wild Card \cite{HannaganProtein2012} &  \cellcolor{gray!25} & \cellcolor{gray!25} & \cellcolor{gray!25} & \cellcolor{gray!25}  &  $0.80$ & \cellcolor{gray!25} \\  \hline         
         Lev/max & Y  & \cellcolor{gray!25} & \cellcolor{gray!25} &  \cellcolor{gray!25} &  $0.74$ & \cellcolor{gray!25} \\  \hline      
         Lev/max & N & \cellcolor{gray!25} & \cellcolor{gray!25} & \cellcolor{gray!25}  &  $0.83$ & \cellcolor{gray!25} \\  \hline      HV & N & 7 & 0 & Simp  &  $0.73$ & $0.005$ \\  \hline     
         
         HV & N & 4 & 1 & Simp  &  $0.81$ & $0.004$ \\  \hline    

         HV & N & 3 & 2 & Simp  &  $0.84$ & $0.004$ \\  \hline             

         HV & N & 3 & 3 & Jac  &  $0.82$ & $0.002$ \\  \hline             
         HV & N & 3 & 4 & Jac  &  $0.82$ & $0.002$ \\  \hline                      
         
         HV & Y & 10 & 0 & Simp  &  $0.65$ & $0.007$ \\  \hline             
         
         HV & Y & 8 & 1 & Simp  &  $0.70$ & $0.006$ \\  \hline            

         HV & Y & 8 & 2 & Simp  &  $0.72$ & $0.006$ \\  \hline

         HV & Y & 5 & 3 & Simp  &  $0.72$ & $0.005$ \\  \hline

         HV & Y & 4 & 4 & Simp  &  $0.73$ & $0.005$ \\  \hline
         
    \end{tabular}
\end{table}

\subsection{Modeling the visual similarity of words}

In~\cite{HannaganProtein2012}, the experimental data on the visual word identification by humans were adapted from~\cite{DavisSpatial2010}; 45 pairs of prime-target strings were obtained, for which there exist the times of human word identification with different types of priming. These pairs do not account for the additional effects, such as frequency and proximity, which were present in the other 16 word pairs in~\cite{DavisSpatial2010}, but are not relevant for assessing models of visual word representations.

Fig.~\ref{fig:human:corr} shows the average value of the Pearson correlation coefficient Corr (between the $\text{sim}_{\text{HV}}$ values and the priming times) vs $R$ for different $s$ ($D = 10000$, mean of $50$ HV realizations). It can be seen that the value of Corr substantially depends on $R$ and $s$, the maximum values were obtained at $R = 3$ and $s = 2$. 
Table~\ref{tab:HannaganProtein2012} shows the Corr values between the times of human identification and the values of similarity. 
The results from~\cite{HannaganProtein2012} for several similarity measures of string vector representations are given. In these measures, a vector is formed for a string with the components corresponding to certain combinations of its letters. Results are presented for: 
\begin{itemize}
    \item Spatial Coding: adapted from~\cite{ DavisSpatial2010}.
    \item GvH UOB: All subsequences of two letters are used, subsequences are not repeated. The similarity is calculated as $|\mathbf{x} \wedge \mathbf{y}| / |\mathbf{y}|$.
    
    \item Kernel UOB (Gappy String kernel): uses counters of all subsequences of two letters within a window. The similarity is determined by $\text{sim}_{\text{cos}}$. 
    
    \item 3-WildCard (gappy kernel): a variant of the kernel similarity of strings proposed in~\cite{HannaganProtein2012}. All subsequences of two letters are padded with $*$ in all acceptable positions. The vector contains the frequency of occurrence for each obtained combination of three symbols, the similarity is determined by $\text{sim}_{\text{cos}}$. For this kernel, the highest Corr value was obtained in~\cite{HannaganProtein2012}. 

\end{itemize}

We also provide the results with the similarity measure 
$1-\text{dist}_{\text{Lev}}(x,y)/{\text{max}}(|x|,|y|)$.

For our HV similarity measures, the average values of Corr, as well as the std values, are given for $50$ HV realizations ($D = 10000$, various $R$ and $s$). The best results among the similarity measures for the particular combination of $R$ and $s$ are demonstrated. It can be seen that with the proper choice of parameters, the results of the HV similarity measures are on a par with other best results, such as $\text{dist}_{\text{Lev}}$, and are slightly better than the results of~\cite{HannaganProtein2012}.

Fig.~\ref{fig:human:data} presents examples of scatterplots with the human priming data ($x$-axis) and similarity ($y$-axis) obtained by some measures of string similarity. The human indicators show the visual identification time by humans for various primes and targets, in milliseconds. The higher the value of the indicator, the earlier the subjects identified the word compared to the neutral primes. 
Each of the $45$ points corresponds to one of the $45$ test string pairs. Corr values are also shown. The scatterplots are depicted for:

\begin{itemize}
    \item top-left, 3-WildCard kernel;
    \item top-right, our HV measure $\text{sim}_{\text{HV}}$, type = cosine; 
    \item bottom-left, our HV measure $\text{sim}_{\text{HV}}$, type = Simpson;
    \item bottom-right, our HV measure $\text{sim}_{\text{HV}}$, type = Jaccard. 
\end{itemize}
For $\text{sim}_{\text{HV}}$, the results were obtained by averaging over $50$ realizations of HVs ($D = 10000$) at $R = 3$, $s = 2$, without the db option. 
Note that for the first task, our best results were obtained with the db option, while for the second task the performance is higher without the db. This discrepancy could be interpreted as dissociation of our model and human perception of string similarity, or suggest various possible modes in human perception of string similarity.

\section{Discussion}
\label{sect:discussion}

\subsection{Related work}
\label{sec:related}

Recently, a detailed account for the related work on representation of sequences was provided in~\cite{RachkovskijEquivariant2021, KleykoSurveyVSA2021Part1}, so here we only sketch a grand schema. 
Largely, the approaches for transforming sequences into HVs are based on either multiplicative or permutative binding~\cite{ SokolovSequence2006}. When a position is assigned with either a random HV (e.g.,~\cite{PlateHolographic2003, HannaganHolographic2011, ImaniDNA2018}, see also~\cite{GallantRepresenting2013} for matrices as positions) or a random permutation (e.g.,~\cite{SahlgrenOrder2008, KanervaHyperdimensional2009}, see also~\cite{PlateHolographic2003}) the resultant representation is either not shift equivariant or not similarity-preserving for the same symbols at nearby positions.

To obtain the equivariance property, it was proposed to use permuted HVs to represent positions~\cite{SahlgrenOrder2008, KanervaHyperdimensional2009}. Also, recursive multiplicative role binding could be used for positions' representation~\cite{ PlateHolographic2003}. However, these options do not allow for similarity preservation of symbols at nearby positions since both the permutation and recursive bindings produce dissimilar HVs even for adjacent positions.

To achieve the similarity-preservation property, it was proposed to use correlated HVs for the representations of nearby positions~\cite{KussulSequences1991, SokolovSequence2006, CohenOrthogonality2013}, see also~\cite{GallantPositional2016} for a matrix version. 
The other option is to use partial correlated permutations~\cite{ KussulPermutation2006,CohenEARP2018}. Both options in their original realization do not provide equivariance to the sequence shifts.

So, the mentioned approaches for representing sequences do not satisfy similarity preservation and shift equivariance simultaneously.
There are, however, two proposals that satisfy both these requirements. 
First, based on the ideas of~\cite{PlateHolographic2003}, in~\cite{FradyDisentangling2018, KomerContinuous2019, VoelkerFPEDynamical2021, FradyFunctions2021, FradyFunctionsNICE2022} an approach using multiplicative binding is considered. It represents a position value by first choosing a complex-valued atomic HV, as in FHRR. This HV is used as the base vector for forming representations of all the positions. Position's value is used to exponentiate each component of the base vector. The HV similarity decreases according to some kernel, where the shape of the kernels depends of the probability distribution used to sample phases. In the case of the uniform distribution, the kernel will approximate the sinc function. This approach is known as fractional power encoding.
The real-valued base HV can be obtained by the inverse FFT of a special complex-valued base vector, i.e., generated so that the phasor entry for a negative frequency is the complex conjugate of the phasor entry for the corresponding positive frequency which is chosen randomly. For the real-valued base HV, binding by circular convolution is used, which is equivalent to the component-wise multiplication in the frequency domain. So, working with real-valued HVs (e.g., within the HRR model) requires computationally expensive forward and inverse FFT and is, therefore, more  costly than just directly using complex HVs in the frequency domain. 
Such a representation of positions could be adapted to the representation of sequences. Similarity preservation can be achieved by setting the bandwidth parameter for the positional HV component exponentiation to be less than one when moving to the next position in the sequence. Mathematically, this representation also ensures equivariance. However, the performance of such a scheme still remains to be explored in practical tasks that require both equivariance and similarity-preserving property, see also~\cite{SchlegelMultivariate2021, VoelkerFPEDynamical2021 }. 

Second, recently, in~\cite{RachkovskijEquivariant2021} a proposal was made that is based on permutations in the SBDR model. Our approach (section III) is based on the same conceptual idea as in~\cite{RachkovskijEquivariant2021}, but instead of permutations it uses recursive multiplicative binding within the FHRR model.

\subsection{Summary and future work }

In this paper, we proposed a new distributed representation of sequences that differs from most of the existing techiques by being both  similarity-preserving for sequence elements at nearby positions and equivariant with respect to sequence shifts. The proposed representation was explored using the tasks of modeling human perception of word similarity by the similarity of symbolic strings, where the HVs of strings were formed as the superposition of their symbols' HVs bound with the HVs of the corresponding positions. 
The considered tasks have demonstrated the importance of similarity-preserving property, as the representations without such a property ($R = 1$) performed poorly on the tasks. This also concerns similarity calculation taking into account sequence shifts, since variants with $s > 0$ performed consistently better than $s = 0$.
The obtained results were either better or on a par with the results of alternative approaches that rely on additional information about sequences, such as n-grams or subsequences.

We reported the results using several hypervector similarity measures such as cosine, Jaccard, and Simpson similarities, but it would be interesting to consider performance using other similarity measures. Another important issue is recovering initial data from their hypervectors, where using the associative 
memory approach~\cite{GritsenkoAMSurvey2017, FrolovWillshaw2002, FrolovTime2006, 
FradyResonator2020, 
KentResonatorNetworks2020,KleykoPrimes2022} adapted to the particular HDC/VSA model is promising.

While the proposed approach was demonstrated with complex-valued HVs within the FHRR model, it could be extended to other HDC/VSA models in which the binding operation is multiplicative and can be applied recursively. This requires that binding an HV with itself should return an HV of the same unit norm without changing the distribution of components' values.
For example, for the HRR model with real-valued HVs and circular convolution as the binding operation, valid real-valued atomic HVs can be obtained with the inverse FFT of the complex-valued unitary HV that satisfies the same constrains as in the real-valued variant of fractional power encoding (see section~\ref{sec:related} and~\cite{HRRLearningNeurips2021}). For the matrix version~\cite{GallantRepresenting2013}, an orthonormal matrix could be considered for the recursive binding.

As a part of the future work, it will also be important to assess how the proposed representation scheme performs in other applications. As examples of such applications we immediately see the ones reported in~\cite{RachkovskijEquivariant2021}: classification of molecular data and spellchecking, as well as other tasks within the areas relying on sequences~\cite{SchlegelMultivariate2021, VoelkerFPEDynamical2021 }.

Finally, while in this paper we have not focused on using the fractional power encoding~\cite{FradyDisentangling2018, KomerContinuous2019, VoelkerFPEDynamical2021, FradyFunctions2021, FradyFunctionsNICE2022}, as we mentioned in section~\ref{sec:related}, given the appropriate scaling of the kernel width parameter it is expected to provide results similar to the proposed approach; hence, we plan to investigate it as part of the future work.

\bibliographystyle{IEEEtran} 
\bibliography{references}

\end{document}